\documentclass{article}
\usepackage{spconf,amsmath, graphicx, array, book tabs, tabularray, multirow}

\usepackage{enumitem}
\setlist{nosep, leftmargin=14pt}

\usepackage{mwe} 
\usepackage{placeins}
\usepackage{opensans}
\usepackage{multirow}
\usepackage{threeparttable}
\usepackage{adjustbox}
\usepackage{xcolor}
\usepackage{colortbl}
\usepackage{lipsum}                 
\usepackage{makecell} 
\usepackage{listings}
\usepackage{makecell}

\definecolor{codegreen}{rgb}{0,0.6,0}
\definecolor{codegray}{rgb}{0.5,0.5,0.5}
\definecolor{codepurple}{rgb}{0.58,0,0.82}
\definecolor{backcolour}{rgb}{0.95,0.95,0.92}

\definecolor{lightgrayd9}{HTML}{e9e9e9}
\definecolor{darkgray96}{HTML}{bdbdbd}

\lstdefinestyle{mystyle}{
    backgroundcolor=\color{backcolour},   
    commentstyle=\color{codegreen},
    keywordstyle=\color{magenta},
    numberstyle=\tiny\color{codegray},
    stringstyle=\color{codepurple},
    basicstyle=\ttfamily\footnotesize,
    breakatwhitespace=false,         
    breaklines=true,                 
    captionpos=b,                    
    keepspaces=true,                 
    numbers=left,                    
    numbersep=5pt,                  
    showspaces=false,                
    showstringspaces=false,
    showtabs=false,                  
    tabsize=2
}

\usepackage{lipsum}

\newcommand{\todo}[1]{}
\renewcommand{\todo}[1]{{\color{red} TODO: {#1}}}

\newcommand{\TODO}[1]{}
\renewcommand{\TODO}[1]{{\color{red} TODO: {#1}}}

\title{MRI Plane Orientation Detection using a Context-Aware 2.5D Model}

\name{SangHyuk Kim$^{1}$ \qquad Daniel Haehn$^{1}$ \qquad Sumientra Rampersad$^{2}$}

\address{
$^{1}$ Department of Computer Science, University of Massachusetts Boston, Boston, Massachusetts, USA \\
$^{2}$Department of Physics, University of Massachusetts Boston, Boston, Massachusetts, USA\
}

\let\OLDthebibliography\thebibliography
\renewcommand\thebibliography[1]{
  \OLDthebibliography{#1}
  \setlength{\parskip}{0pt}
  \setlength{\itemsep}{0pt plus 0.3ex}
}

\begin{document}
%
\maketitle
\begin{abstract}
Humans can easily identify anatomical planes (axial, coronal, and sagittal) on a 2D MRI slice, but automated systems struggle with this task. Missing plane orientation metadata can complicate analysis, increase domain shift when merging heterogeneous datasets, and reduce accuracy of diagnostic classifiers. This study develops a classifier that accurately generates plane orientation metadata. We adopt a 2.5D context-aware model that leverages multi-slice information to avoid ambiguity from isolated slices and enable robust feature learning. We train the 2.5D model on both 3D slice sequences and static 2D images. While our 2D reference model achieves 98.74\% accuracy, our 2.5D method raises this to 99.49\%, reducing errors by 60\%, highlighting the importance of 2.5D context. We validate the utility of our generated metadata in a brain tumor detection task. A gated strategy selectively uses metadata-enhanced predictions based on uncertainty scores, boosting accuracy from 97.0\% with an image-only model to 98.0\%, reducing misdiagnoses by 33.3\%. We integrate our plane orientation model into an interactive web application\footnote{\label{footnote:webinterface}Web interface, https://shkimmie-umb.github.io/plane-classifier-app/} and provide it open-source\footnote{GitHub repository, https://mpsych.org/mri-plane}.
\end{abstract}

%

\section{Introduction}
\label{sec:intro}
\begin{figure*}[hbtp]
    \centering
    \includegraphics[width=2\columnwidth]{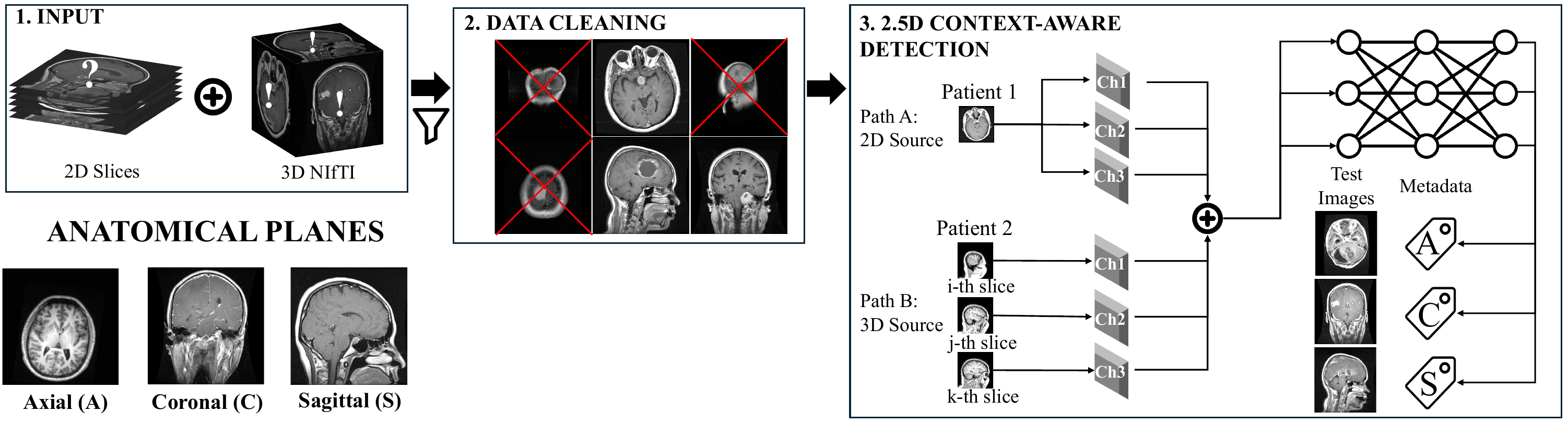} 
    \caption{
    System overview of the 2.5D context-aware plane detection pipeline. (1) Heterogeneous data from 2D-native and 3D-native sources are aggregated and (2) passed through a data cleaning pipeline. (3) The 2.5D model learns a unified strategy, processing 2D images with a static context and 3D images with a random context. (4) The final trained classifier generates the missing plane orientation metadata for any input slice.
    }
    \label{fig:overview}
\end{figure*}

Medical imaging data is processed as 2D slices viewed in axial, sagittal, or coronal planes (see Fig.~\ref{fig:overview}). Labeling each slice with its orientation can enhance feature learning and accuracy for machine learning-based tasks like tumor detection~\cite{helpstumor, dldicom}. However, many 2D MRI datasets lack this data, and 3D NIfTI headers can become corrupted during handling~\cite{corruption1, dldicom}. This limitation discourages large-scale integration of diverse datasets and exacerbates domain-shift issues~\cite{domainshift1, domainshift2}.
While human radiologists easily identify slice planes, machine learning methods can struggle. Prior work often focuses on narrow scopes or tasks. For example, some methods extract only the midsagittal plane~\cite{planeorientation2}.
Pure plane orientation classifiers~\cite{planeorientation1} struggle with ambiguous, isolated slices when looking at a sole image, with authors noting failures on ``near-skull'' images that lack anatomical context. To address these limitations, we present an automated plane orientation pipeline achieving over 99\% accuracy. We investigate a multi-slice random sampling method \texttt{[i, j, k]} using a 2.5D context-aware model~\cite{2.5D_1, 2.5D_2, 2.5D_3} to resolve the ambiguity of isolated 2D slices. We validate the metadata's value in a brain tumor detection task, showing enhanced classifier performance. We adopt a gated strategy leveraging classifier uncertainty to selectively apply metadata, reducing misdiagnoses. Finally, we demonstrate this system in an open-source interactive web application, enabling integration into pipelines for various real-world use cases.

\section{Automatic Plane Detection}
\label{sec:planeorientation}

\subsection{Model Design for Contextual Plane Detection}
\label{subsec:2_modeldesign}
We evaluate our input strategies using two backbone architectures: AlexNet~\cite{alexnet} and ResNet-18~\cite{resnet}. AlexNet (8-layer, 60M params) was chosen for its larger kernels, which excel at capturing global shapes and anatomical landmarks. In contrast, ResNet-18 (18-layer, 11.7M params) was chosen for its residual connections, which enable it to capture fine details and complex spatial relationships. This comparison assesses whether AlexNet’s broad feature extraction or ResNet-18’s deep residual learning better supports the recognition of anatomical planes.
To investigate the impact of contextual information, we design three 3-channel input strategies:


\begin{itemize}
    \item 2D Model: A single slice duplicated (\texttt{[i, i, i]}), serving as a reference without contextual information.
    \item Sequential 2.5D Model: This model uses adjacent slices (\texttt{[i-1, i, i+1]}) to learn sequential local anatomical flow, testing the value of anatomically ordered context.
    \item Random 2.5D Model: Three random slices from the same volume and plane (\texttt{[i, j, k]}), hypothesized to learn robust, identity-level features and act as an effective regularizer.
\end{itemize}

\noindent
\subsection{Validation via Downstream Task}
\label{subsec:2_validationtumor}
We validate inferred orientation metadata from our plane classifier in a four-class tumor classification task: Glioma, Meningioma, Pituitary, and No-Tumor. We establish two models: an Image-Only model using only the 2D image input and a Metadata-Enhanced model integrating a one-hot vector of the plane orientation with 2D image features inferred by our automatic plane detection pipeline.
Applying generated metadata carelessly can harm the tumor classifier's performance by introducing incorrect plane information. To address this, we develop a gated metadata strategy that selects between the two models, relying on the tumor classifier's uncertainty score from predictive entropy~\cite{shannon}. We find the optimal gating threshold to maximize accuracy on the validation set in Fig.~\ref{fig:optimalthreshold} and apply it to the test set. The Metadata-Enhanced model's confident predictions are used; if uncertain, we default to the Image-Only model's prediction. This approach leverages accurate metadata while avoiding errors from incorrect metadata.

\subsection{Web Deployment}
\label{subsec:2_webdeployment}
To make our method directly accessible to real-world users, our final classifier is deployed in an interactive web application. The model is converted to TensorFlow.js format~\cite{smilkov2019tensorflow} for client-side inference in the user's browser. This serverless edge-computing approach requires no server computation and ensures patient privacy, as no medical data leaves the local machine~\cite{kim2025melanoma}. The application provides the predicted plane orientation and an uncertainty score in real time.

\section{Experimental Setup}
\label{sec:experiments}

\subsection{Datasets and Preprocessing}
\label{subsec:datasets}
We utilize the 2D-native BRISC dataset~\cite{brisc}, which consists of full-head T1-weighted MR images of tumor patients, and the 3D-native IXI dataset~\cite{ixi}, which consists of full-head MR images of healthy adults. We use only the T1-weighted images from IXI to align with BRISC. Our pipeline converts 3D NIfTI volumes into clean 2D images through several steps: 1) Apply a 3D morphological opening operation, erosion followed by dilation, to remove noise; 2) Sparse sampling selects every 10th slice to cut redundancy; 3) Discard slices with minimal brain tissue using a threshold of mean intensity $> 0.1$ and voxel coverage $> 0.25$; 4) Pad rectangular slices to a square aspect ratio before resizing to prevent distortion. This filtering results in 0.88\% (53/6000) of BRISC images discarded and 93.81\% (360,488/384,266) of IXI, leaving 23,778 quality slices. For plane classification experiments, we create three balanced training sets with 4,000 samples each through random sampling: BRISC-only, IXI-only, and a BRISC+IXI mixed set (2k/2k). For testing, we use cleaned BRISC (N=1187) and IXI (N=4766) sets. For tumor detection validation, we train models on a combined dataset~\cite{combined} that includes BRISC and IXI and evaluate only on the BRISC test set, as IXI does not contain tumors.

\subsection{Training Strategy}
\label{subsec:3_trainingstrategy}
For our primary classification models, we train AlexNet~\cite{alexnet} and ResNet-18~\cite{resnet} on the Brisc+IXI mixed dataset for 30 epochs with the Adam optimizer~\cite{adamoptimizer}, a learning rate of 0.001, and cross-entropy loss. These models utilize ImageNet~\cite{deng2009imagenet} pre-trained weights and specific normalization~\cite{deng2009imagenet,alexnet}. We apply a data augmentation with random horizontal and vertical flips and rotations up to 20 degrees to enhance robustness, particularly for variant coronal images. To tackle class imbalance in the training set, we adjust sampling probabilities for balanced distribution of axial, coronal, and sagittal slices within each training batch, preventing bias toward overrepresented classes. For downstream tumor detection validation, we build three comparative models—the Image-Only, Metadata-Enhanced, and final Gated-Metadata model—using an identical ResNet-18 backbone. This ensures performance differences stem solely from metadata use, allowing clear comparison of strategies.

\begin{table}[h!]
\caption{A step-by-step comparison identifies the optimal model architecture and training configuration.}
\centering
\label{table:performance_table}
\begin{adjustbox}{width=1\columnwidth}{
\begin{threeparttable} 
\normalsize
\begin{tabular}{l l l l c c c} 
\toprule
\multirow{2}{*}{\textbf{Step}} & \multirow{2}{*}{\textbf{Model}} & \multirow{2}{*}{\textbf{\makecell[l]{Norm.\\Config.}}} & \multirow{2}{*}{\textbf{Train Set}} & \multicolumn{3}{c}{\textbf{Test Set Accuracy (\%)}} \\
\cmidrule(l){5-7} 
& & & & {\textbf{BRISC} (Path.)} & {\textbf{IXI} (Healthy)} & {\textbf{Average}} \\
\midrule

\multirow{3}{*}{\makecell[l]{1. Select Best\\Train Set}} 
& AlexNet\tnote{a}~\cite{alexnet} & MinMax & BRISC-Only~\cite{brisc} & 97.81 & 80.84 & 89.33 \\
& AlexNet\tnote{a} & MinMax & IXI-Only~\cite{ixi} & 57.54 & 98.97 & 78.26 \\
& {\cellcolor{lightgrayd9}AlexNet\tnote{a}} & {\cellcolor{lightgrayd9}MinMax} & {\cellcolor{lightgrayd9}\textbf{BRISC+IXI}} & {\cellcolor{lightgrayd9}97.73} & {\cellcolor{lightgrayd9}98.45} & {\cellcolor{lightgrayd9}\textbf{98.09}} \\
\cmidrule(l){1-7} 

\multirow{2}{*}{\makecell[l]{2. Select Best\\Model}} 
& ResNet18\tnote{a}~\cite{resnet} & MinMax & BRISC+IXI & 95.28 & 98.95 & 97.12 \\
& {\cellcolor{lightgrayd9}\textbf{AlexNet}\tnote{a}} & {\cellcolor{lightgrayd9}MinMax} & {\cellcolor{lightgrayd9}BRISC+IXI} & {\cellcolor{lightgrayd9}97.73} & {\cellcolor{lightgrayd9}98.45} & {\cellcolor{lightgrayd9}\textbf{98.09}} \\
\cmidrule(l){1-7} 

\multirow{2}{*}{\makecell[l]{3. Select Best\\Normalization}}
& {\cellcolor{lightgrayd9}AlexNet\tnote{a}} & {\cellcolor{lightgrayd9}MinMax} & {\cellcolor{lightgrayd9}BRISC+IXI} & {\cellcolor{lightgrayd9}97.73} & {\cellcolor{lightgrayd9}98.45} & {\cellcolor{lightgrayd9}98.09} \\

& {\cellcolor{darkgray96}AlexNet\tnote{a}} & {\cellcolor{darkgray96}\textbf{ImageNet}}  & {\cellcolor{darkgray96}BRISC+IXI} & {\cellcolor{darkgray96}\textbf{98.74}} & {\cellcolor{darkgray96}\textbf{99.79}} & {\cellcolor{darkgray96}\textbf{99.27}} \\
\cmidrule(l){1-7} 
\multirow{2}{*}{\makecell[l]{4. Add Context\\Slices}}
& AlexNet\tnote{b} & ImageNet & BRISC+IXI & 99.49 & 99.98 & 99.74 \\
& {\cellcolor{green!15}\textbf{AlexNet}\tnote{c}} & {\cellcolor{green!15}\textbf{ImageNet}} & {\cellcolor{green!15}\textbf{BRISC+IXI}} & {\cellcolor{green!15}\textbf{99.49}} & {\cellcolor{green!15}\textbf{99.99}} & {\cellcolor{green!15}\textbf{99.74}} \\

\bottomrule
\end{tabular}

\begin{tablenotes}
\item a: 2D model
\hspace{1em} 
b: 2.5D model (sequential slices) \hspace{1em} c: 2.5D model (random slices)
\end{tablenotes}

\end{threeparttable}
}\end{adjustbox}
\end{table}

\section{Results and Discussion}
\label{sec:results}

\subsection{Plane Orientation Model Optimization}
\label{subsec:4_planeorientation}

The optimization process for the plane orientation classifier is in Table~\ref{table:performance_table}. Results show models trained solely on single data sources like Brisc-only or IXI-only face domain-shift issues and lack generalization. The Brisc+IXI mixed dataset achieves the highest average accuracy of 98.09\%, with AlexNet surpassing ResNet18 at 97.12\%. Using ImageNet pre-training reaches 99.27\% accuracy, compared to 98.09\% when training from scratch. Thus, we select the pre-trained AlexNet on Brisc+IXI as the optimal 2D model. We compare this 2D reference with our 2.5D contextual models, where both 2.5D sequential and random contextual models achieve a higher average accuracy of 99.74\%. The 2.5D random contextual model achieves the peak accuracy of 99.99\% on the IXI dataset, slightly outperforming the 2.5D sequential model as seen in Table~\ref{table:performance_table} Step 4. The 2.5D random model reduces misclassifications by 60\% from 15 to 6 compared to the 2D-only model. This confirms that using 2.5D contextual information is crucial for resolving ambiguous slices, leading us to select the 2.5D random contextual model as our final architecture for subsequent tasks.

\subsection{Comparative Error Analysis: 2D vs. 2.5D Models}
\label{subsec:4_interpretability}
\begin{figure}[h!]
\centering
\includegraphics[width=1\columnwidth]{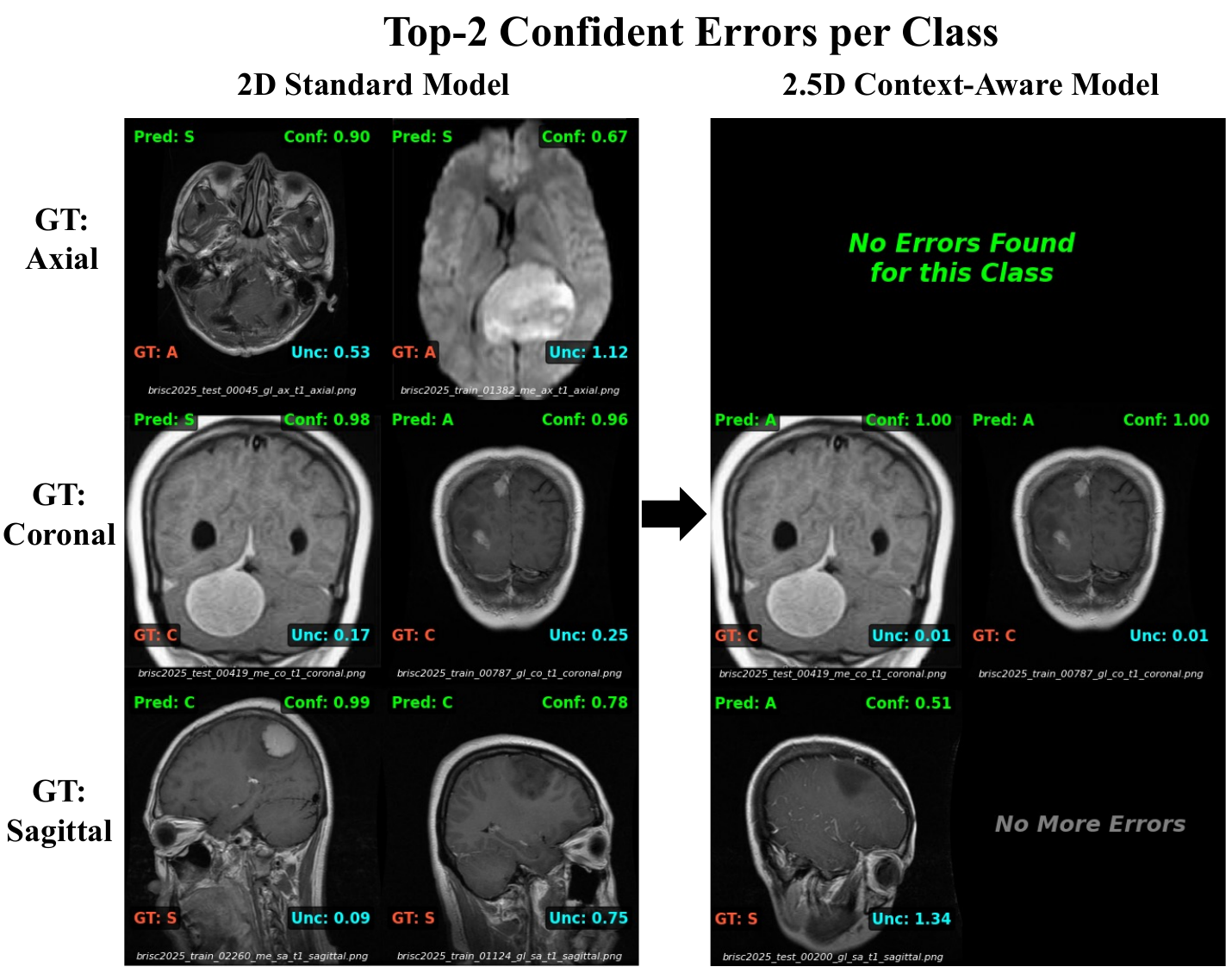}
\caption{Comparing 2D vs. 2.5D errors on the BRISC test set~\cite{brisc}. The 2.5D model (right) corrects most of the 2D model's top misclassifications (left).}
\label{fig:detectivework}
\end{figure}

To qualitatively demonstrate the impact of the 2.5D model, we visualize its most confident misclassifications alongside those of the 2D reference model using the BRISC test set as shown in Fig.~\ref{fig:detectivework}. This analysis highlights the 2.5D model's ability to resolve ambiguity through context. The 2.5D model eliminates all errors from the 2D model in the Axial class. Notably, in the Coronal class (GT: `C'), both models' most confident error is the same image; however, the 2D model labels it as sagittal ('S'), while the 2.5D model labels it as axial ('A'). The image features a large tumor causing asymmetry, misleading the 2D model to incorrectly label the slice as sagittal. In contrast, the 2.5D model disregards this asymmetry and identifies the slice as axial, which, although still flawed, is more coherent in this context. In the Sagittal class, the 2D model's two most confident errors get corrected by the 2.5D model, leaving only one error.

\subsection{Validating Metadata Utility: Brain Tumor Detection}
\label{subsec:4_tumordetection}

\begin{figure}[h!]
\centering
\includegraphics[width=1\columnwidth]{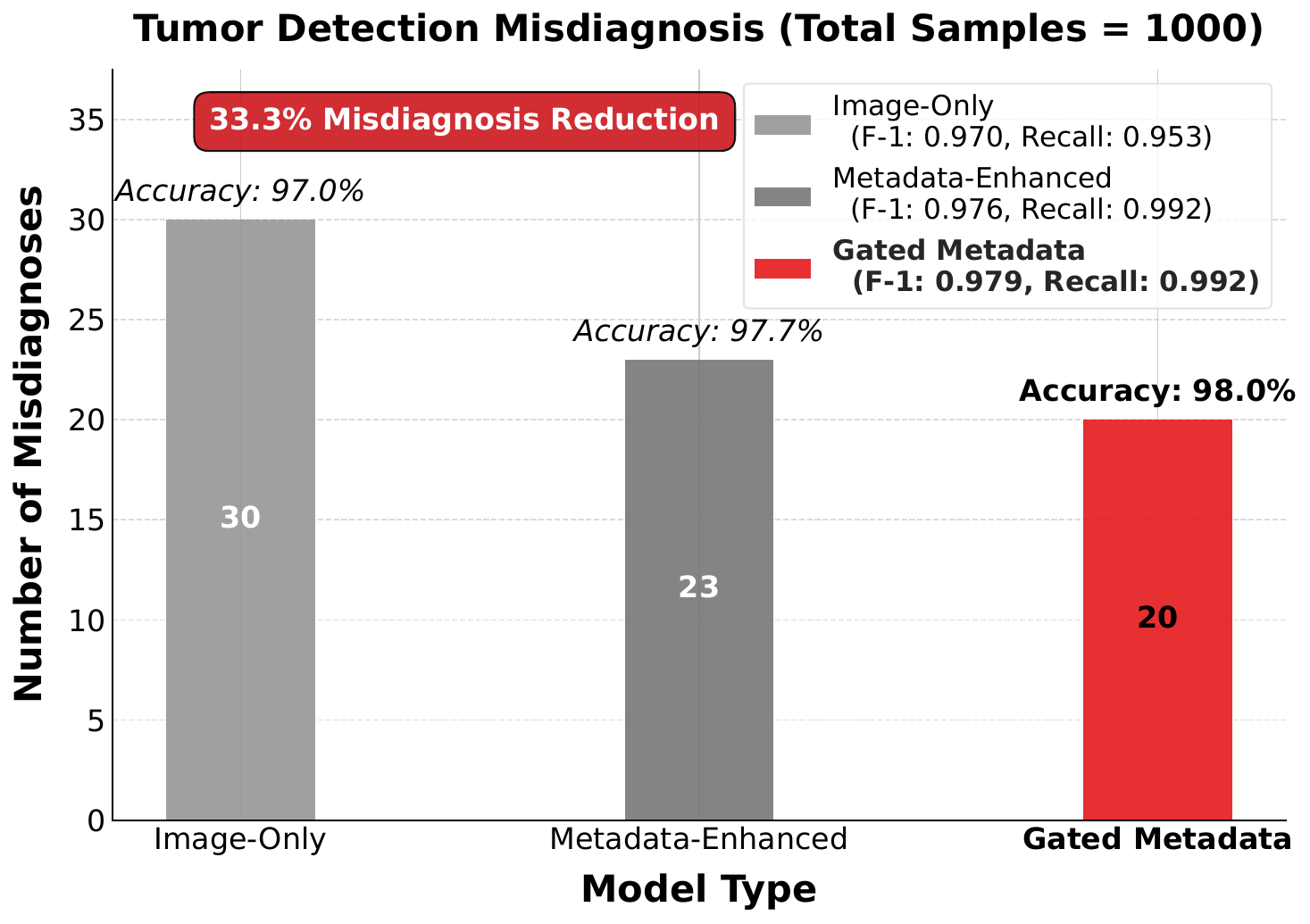}
\caption{Tumor Detection performance comparison. The Gated Metadata model achieves the highest accuracy (98.0\%) and the fewest misdiagnoses (20), representing a 33.3\% reduction in errors compared to the Image-Only model.}
\label{fig:misdiagnosis_red}
\end{figure}

To assess the practical value of our generated plane metadata, we apply it to a downstream brain tumor detection task, summarized in Fig.~\ref{fig:misdiagnosis_red}. Our Image-Only model achieves an initial accuracy of 97.0\%, resulting in 30 misdiagnoses on the BRISC test set. Adding plane metadata in an ``always-on'' manner (Metadata-Enhanced model) improves accuracy to 97.7\% and reduces misdiagnoses to 23. Our final Gated Metadata model, which selectively applies metadata based on classifier confidence, achieves the best performance. 

\begin{figure}[h!]
\centering
\includegraphics[width=0.75\columnwidth]{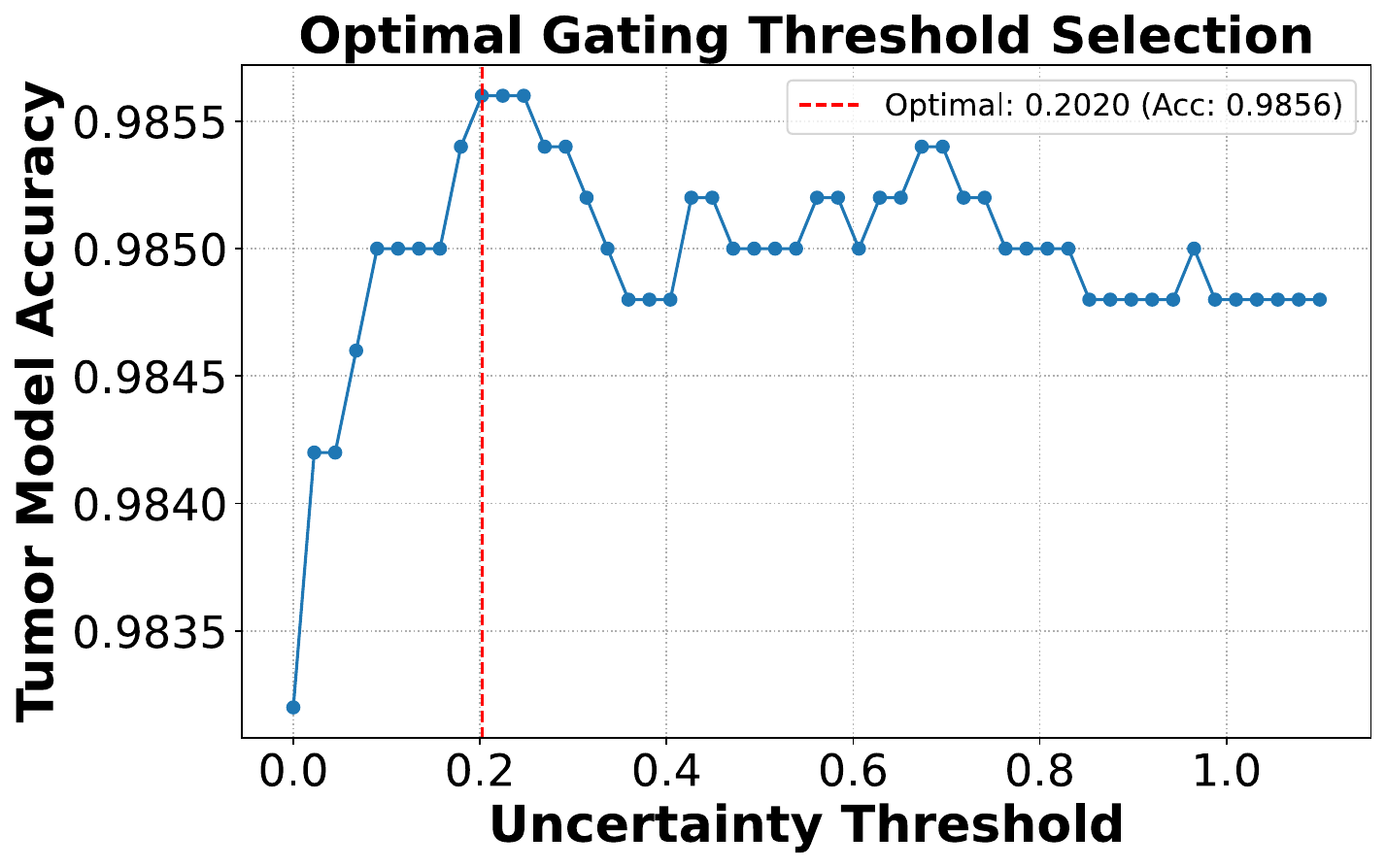}
\caption{Gated model accuracy plotted against the uncertainty threshold on the validation set.}
\label{fig:optimalthreshold}
\end{figure}

To find the best fusion strategy, we plotted gated model accuracy against the uncertainty threshold on the validation set (Fig.~\ref{fig:optimalthreshold}). This threshold determines when to use the Image-Only versus the Metadata-Enhanced prediction. The optimal threshold of 0.2020 (vertical line) yielded the maximum accuracy of 98.56\% and was subsequently used for the test set evaluation.

\begin{figure}[h!]
\centering
\includegraphics[width=0.8\columnwidth]{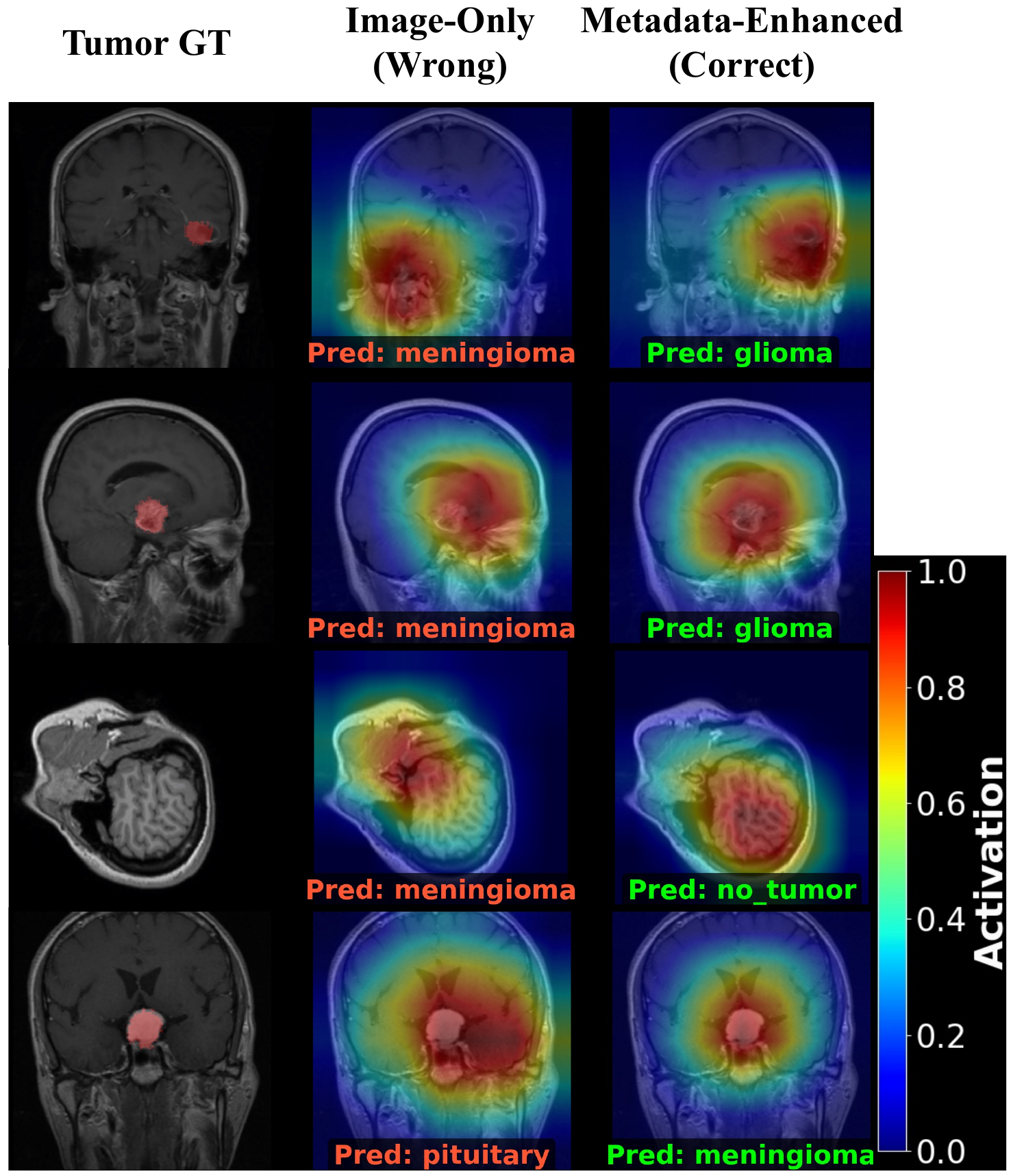}
\caption{Qualitative analysis of tumor detection using Grad-CAM~\cite{gradcam}. The first column shows the location of the tumor (red) if one is present.}
\label{fig:gradcam}
\end{figure}

The gated model achieves 98.0\% accuracy, lowering total misdiagnoses to 20. This represents a 33.3\% reduction in misdiagnoses, correcting 10 errors compared to the Image-Only model, indicating the practical value of our generated metadata.
Qualitative analysis using Grad-CAM~\cite{gradcam} (Fig.~\ref{fig:gradcam}) supports this quantitative result. These visualizations reveal how plane metadata rectifies some tumor detection errors. For instance, the Image-Only model misclassifies a Glioma slice due to incorrect attention (first row), whereas the Metadata-Enhanced model correctly targets and predicts the tumor. On a No-Tumor slice (third row), the Image-Only model generates a false positive because it focuses on an area outside of the brain, whereas the Metadata-Enhanced model focuses on the brain and classifies the image as negative.

\subsection{Ablation Study}
\label{subsec:4_ablation}
To evaluate different 2.5D contextual strategies for plane orientation detection, we conduct an ablation study. To isolate the effect of context and rigorously test generalization, models in Table \ref{tab:ablation} train from scratch on the IXI-only dataset, which contains sequential information, and test on the unseen 2D non-sequential BRISC dataset. This setup removes the benefits of pre-training and uses a single training domain, resulting in lower accuracies than our main models while demonstrating the impact of contextual information. Results show both the random (63.5\%) and sequential (73.0\%) 2.5D models with 3 context slices achieve higher accuracy than the 2D reference models (48.7\% no-aug, 51.9\% full-aug). This further suggests that in this isolated training scenario, local anatomical flow is vital and acts as a key feature.

\begin{table}[h!]
\centering
\caption{
    Ablation study of 2.5D context strategies.
    All models are AlexNet, trained from scratch on the IXI-only dataset~\cite{ixi}
    with min-max (0-1) normalization.
    Accuracy is reported on the unseen BRISC~\cite{brisc} test set.
}
\label{tab:ablation}
\begin{adjustbox}{width=1\columnwidth}{
\begin{tabular}{l c c l c}
\toprule
\textbf{Step} & \textbf{Model Type} & \textbf{\makecell[l]{Context\\Slices (N)}} & \textbf{Context Strategy} & {\textbf{Accuracy (\%)}} \\
\midrule

\multirow{2}{*}{\makecell[l]{1. Establish 2D\\Reference}}
& 2D & 1 & No Augmentation & 48.7 \\
& 2D & 1 & Full Augmentation & 51.9 \\
\midrule

\multirow{2}{*}{\makecell[l]{2. Investigate\\Slice Count}}
& 2.5D & 2 & Random \texttt{[i, j]} & 52.7 \\
& 2.5D & 3 & Random \texttt{[i, j, k]} & 63.5 \\
\midrule

\makecell[l]{3. Select Optimal\\2.5D Strategy}
& \textbf{2.5D} & \textbf{3} & \textbf{Sequential \texttt{[i-1, i, i+1]}} & {\textbf{73.0}} \\

\bottomrule
\end{tabular}
}\end{adjustbox}
\end{table}

\section{Conclusions}
\label{sec:conclusion}
We develop and validate a 2.5D context-aware classifier to generate missing MRI plane orientation metadata. Our 2.5D random-sampling model, trained on a mixed-domain dataset, proves most effective, reducing misclassifications compared to a 2D reference model. We demonstrate the practical value of this generated metadata in a tumor detection task, where our gated strategy based on classifier confidence decreases errors by 33.3\% over the image-only model. An ablation study finds that local anatomical flow is another key feature. Future work will expand this contextual framework to other metadata, such as T1 vs T2-weighted scans.

\section{Acknowledgments}
This research was funded in part
by the Massachusetts Life Sciences Center through grant
Bits-to-Bytes 34428 and the National Institutes of Health through grant R01NS133229.

\section{Compliance with ethical standards}
\label{sec:print}

All datasets used in this study were anonymized and publicly available, for which no ethical approval was required.

\bibliographystyle{IEEEbib}
\bibliography{strings,refs}

\end{document}